\pdfoutput=1

\documentclass[11pt]{article}

\usepackage[]{naacl2021}

\usepackage{times}
\usepackage{latexsym}

\usepackage[T1]{fontenc}

\usepackage[utf8]{inputenc}

\usepackage{microtype}

\usepackage{graphicx}
\usepackage{multirow}
\usepackage{booktabs}
\usepackage{amsmath}

\def\eg{\emph{e.g}.~}
\def\ie{\emph{i.e}.~}

\title{Improving Text-to-SQL with Schema Dependency Learning}

\small
\author{
  Binyuan Hui \thanks{\quad Equal contribution.}, Xiang Shi\footnotemark[1], Ruiying Geng, Binhua Li, Yongbin Li \thanks{ \quad Corresponding author.}, Jian Sun\\
  Alibaba Group \\ 
  \small
  \texttt{binyuan.hby@alibaba-inc.com, sxron.sx@alibaba-inc.com}\AND
  Xiaodan Zhu \\
  Ingenuity Labs Research Institute \& ECE, Queen’s University \\
  \small
  \texttt{zhu2048@gmail.com}
}

\begin{document}
\maketitle
\begin{abstract}

Text-to-SQL aims to map natural language questions to SQL queries. The sketch-based method combined with execution-guided (EG) decoding strategy has shown a strong performance on the WikiSQL benchmark. However, execution-guided decoding relies on  database execution, which significantly slows down the inference process and is hence unsatisfactory for many real-world applications. In this paper, we present the Schema Dependency guided multi-task Text-to-SQL model (\texttt{SDSQL}) to guide the network to effectively capture the interactions between questions and schemas. The proposed model outperforms all existing methods in both the settings with or without EG. We show the schema dependency learning partially cover the benefit from EG and alleviates the need for it. \texttt{SDSQL} without EG significantly reduces time consumption during inference, sacrificing only a small amount of performance and provides more flexibility for downstream applications.

\end{abstract}

\section{Introduction}
Text-to-SQL is a sub-area of semantic parsing that has received an intensive study recently \citep{DBLP:conf/aaai/ZelleM96,DBLP:conf/uai/ZettlemoyerC05,DBLP:conf/acl/WongM07,DBLP:conf/emnlp/ZettlemoyerC07,Li2014ConstructingAI,Yaghmazadeh2017SQLizerQS,DBLP:conf/acl/IyerKCKZ17}. It aims to translate a nature language question to an executable SQL.
This task underlies many applications such as table-based question answering and fact verification \citep{chen2019tabfact, Herzig2020TAPASWS,yang-etal-2020-program}.
Recently, complex Text-to-SQL settings have been proposed, \eg, Spider \citep{DBLP:conf/emnlp/YuZYYWLMLYRZR18}, SparC \citep{DBLP:conf/acl/YuZYTLLELPCJDPS19} and CoSQL \citep{DBLP:conf/emnlp/YuZELXPLTSLJYSC19}.
However, generating SQL for individual queries in the single table setup (WikiSQL \citep{Zhong2017Seq2SQLGS}) is still the most fundamental.
The existing single table Text-to-SQL models can be divided into two categories: \textit{generation-based} or \textit{sketch-based} models.
The generation-based methods \citep{Dong2016LanguageTL, Krishnamurthy2017NeuralSP, Sun2018SemanticPW, Zhong2017Seq2SQLGS} decode SQL based on a sequence-to-sequence process, mainly using the attention and copying mechanism.
Such models suffer from the \textit{ordering issue} when generating SQL sequences since they do not sufficiently enforce SQL syntax. 
The sketch-based methods \cite{Xu2017SQLNetGS, Dong2018CoarsetoFineDF, yu-etal-2018-typesql,Hwang2019ACE, He2019XSQLRS} have been further proposed to avoid the \textit{ordering issue} and achieve better performances on the WikiSQL benchmark.
The models decompose SQL generation procedure into sub-modules, \eg, \textit{SELECT column, AGG function, WHERE value}, etc.

In addition, the execution-guided (EG)  decoding strategy \citep{Wang2018RobustTG} runs the acquired SQL queries. The outcome (\eg, whether the database engine returns run-time errors) can be used to guide Text-to-SQL.
Although EG can significantly improve Text-to-SQL performance, it depends on the SQL execution over databases, which greatly impairs the speed of inference and hence practical applications.
The run-time errors are mainly caused by mismatches between the generated components and operators (\eg, sum over a column with the \textit{string} type). The correctness of these components can depend on schema linking, i.e., the interaction between schemas and questions. 
Hence effectively modelling this interaction is a feasible way to circumvent the EG requirement.

\begin{figure*}
	\centering
	\includegraphics[width=1\linewidth]{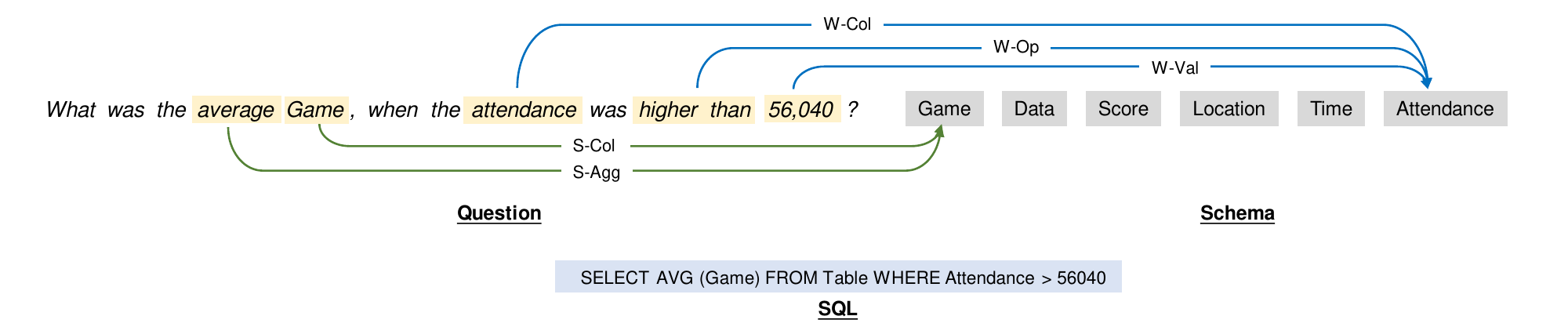}
	\caption{An example of schema dependency learning.}
	\label{sd}
\end{figure*}

To this end, we propose a novel model based on \textit{schema dependency}, which is designed to more effectively capture the complex interaction between questions and schemas.
Our proposed model is called \textbf{S}chema \textbf{D}ependency guided multi-task Text-to-\textbf{SQL} model (\texttt{SDSQL}). Our model aims to integrate schema dependency and SQL prediction simultaneously and adopts an adaptive multi-task loss for optimization.
Experiments on the WikiSQL benchmark show that \texttt{SDSQL} outperforms existing models. Particularly in the setup without the EG strategy, our model significantly outperforms the existing models.


\section{Related Work}
Previous models \citep{Dong2016LanguageTL, Krishnamurthy2017NeuralSP, Sun2018SemanticPW, Zhong2017Seq2SQLGS} leverage Seq2Seq models to translate questions to SQL in the single table setup, which is called generation-based method.
The sketch-based method achieve better performance, firstly SQLNet \citep{Zhong2017Seq2SQLGS} decomposes SQL into several independent sub-modules and perform classification.
Base on that, TypeSQL \citep{Yu2018TypeSQLKT} introduces the type information to better understand rare entities in the input. The Coarse-to-Fine model \citep{Dong2018CoarsetoFineDF} performs progressive decoding.
Further more, SQLova \cite{Hwang2019ACE} and X-SQL \cite{He2019XSQLRS} utilize pre-trained language models in encoder and leverage contextualization to significantly improve performance.
IE-SQL \citep{Ma2020MentionEA} proposes an information extraction approach to Text-to-SQL and tackles the task via sequence-labeling-based relation extraction.
To tackle the run-time error, \citet{Wang2018RobustTG} takes the execution-guided (EG) strategy, which further improves the performance.
More recently, the advantage of schema linking in semantic parsing has also been explored by \citep{guo-etal-2019-towards} and \citep{wang-etal-2020-rat}, \ie, the former uses heuristic rules to identify the question-schema relation and feeds this information as input while learning the representation of the question; the latter work formulates a question-contextualized schema graph and encodes the question-schema interaction via attention.
Compared to them, \texttt{SDSQL} benefits from more specific SQL-related linking types (e.g. S-Col, S-Agg), which depends on the corresponding SQL for linking generation. Also, the proposed dependency approach is more explicit and logical.

\section{The Model}
\begin{figure}
	\centering
	\includegraphics[width=0.9\linewidth]{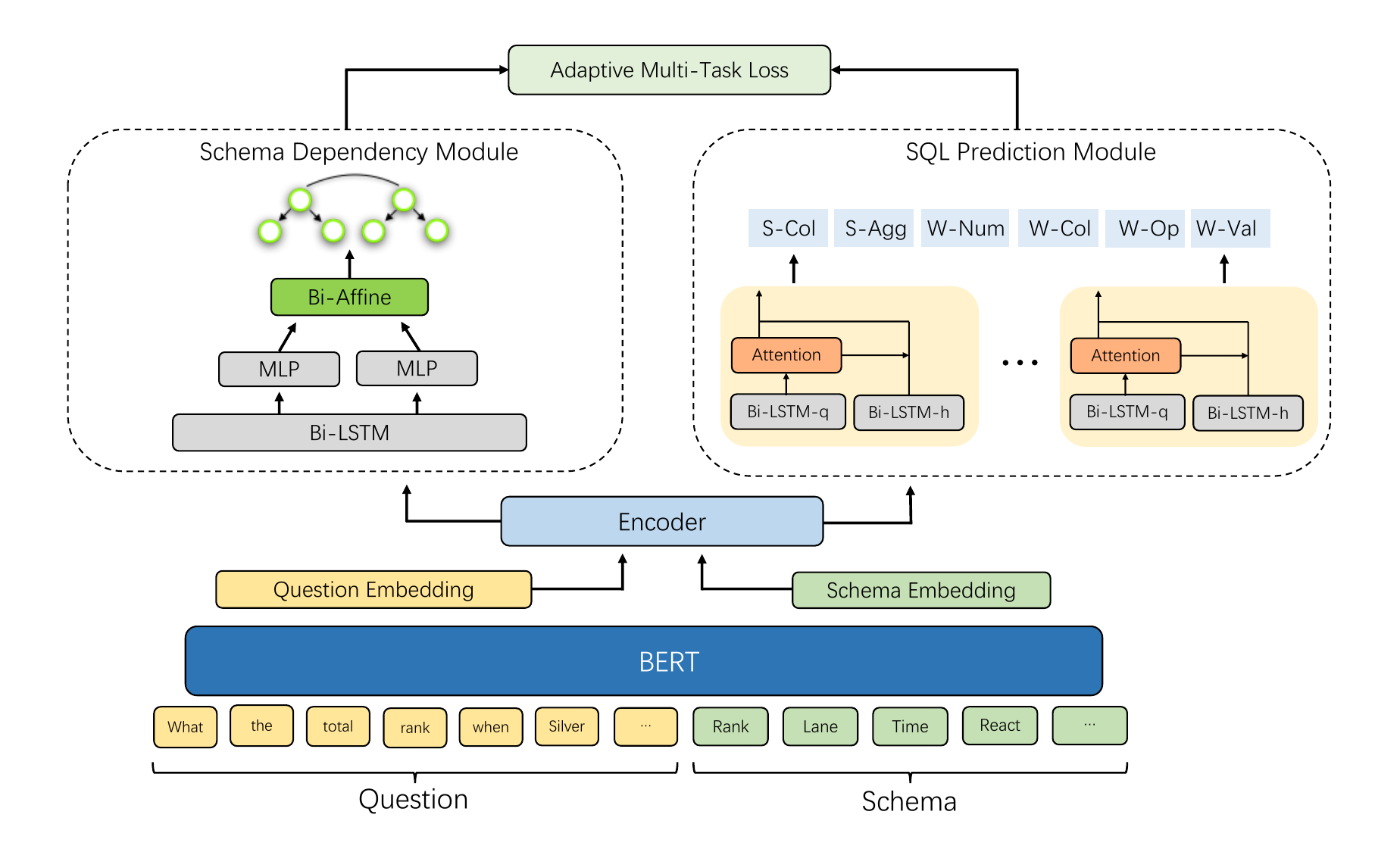}
	\caption{Illustration of the \texttt{SDSQL} model architecture.}
	\label{net}
\end{figure}

The overall architecture of the proposed \texttt{SDSQL} model is depicted in Figure \ref{net}.

\subsection{Input Representation}
We denote a natural language question as $Q = \left\langle{q}_{1}, \ldots, {q}_{n}\right\rangle$, where $n$ is the length and $q_i$ the {$i$-th} token.
The headers of the schema involved in the question can be expressed as $S = \left\langle{s}_{1}^{1}, {s}_{1}^{2}, \ldots, {s}_{m}^{1}, {s}_{m}^{2}, \ldots\right\rangle$, where $s_i^{j}$ denotes the {$j$-th} token of the {$i$-th} header, and $m$ is the total number of headers.
We adopt BERT \cite{Devlin2019BERTPO} to encode question $Q$ and headers $S$:

\begin{equation}\label{input}
\resizebox{.88\hsize}{!}{$
\mathtt{\texttt{[CLS]}}, q_{1},  \ldots , q_{n}, \mathtt{\texttt{[SEP]}}, s_{1}^{1}, s_{1}^{2} , \ldots, \mathtt{\texttt{[SEP]}}, s_{m}^{1}, s_{m}^2 \ldots \mathtt{\texttt{[SEP]}}.
$}
\end{equation}

\paragraph{Encoder.} The output embedding from BERT is fed to a two-layer Bi-LSTM encoder to obtain task-related representation.
For clarity, we denote the encoder output of the {$i$-th} token of the question as $x_i$.
The tokens of each individual header are fed to the encoder separately. $h_l$ denotes the output of the final embedding for the {$l$-th} header. 

\subsection{Schema Dependency Learning}
\paragraph{Data Construction}
In order to capture the explicit and complex interaction between questions and headers, we propose the schema dependency learning task.
Given a question and its corresponding SQL statement, we use alignment to construct dependency data between question tokens and headers.
Specially, we pre-define a series of dependency labels: \textit{select-column (S-Col)}, \textit{select-aggregation (S-Agg)}, \textit{where-column (W-Col)}, \textit{where-operator (W-Op)} and \textit{where-value (W-Val)}.
For training labels, we use the automatic annotation method to generate the linking relationships between schemas and questions using the corresponding SQL labels. 
As shown in Figure \ref{sd},
for example, given a question “What was the average Game, when the attendance was higher than 56,040?” and schema “\texttt{[Game], [Data], [Source], [Location], [Time], [Attendance]}”, the corresponding SQL should be “SELECT AVG (Game) FROM Table WHERE Attendance > 56040”. Guided by elements mentioned in the SQL, we extract the mentioned column in clause, and get corresponding token spans in the question that are logically related to the column with heuristic rules (e.g. n-gram, stemming):
\begin{itemize}
    \item “AVG(Game)” in SQL guides the link of “Game” in question and \texttt{[Game]} in schema with S-col label, and “average” in question and \texttt{[Game]} in schema with S-agg label.
    \item “Attendance” in SQL WHERE clause guides the link of “attendance” in question and \texttt{[Attendance]} in schema with W-col label.
    \item “56040” in SQL matches the “56,040” in question, W-val label could be added between “56,040” and \texttt{[Attendance]} on the basis of W-col label.
    \item for operator , we pre-define the possible description style of the operator “>”, e.g. “rather than, higher than, bigger than, larger than”. Then match the question to build a W-op label.
\end{itemize}
The goal of schema dependency is to find dependency label between the tokens of question and schema.

\paragraph{Dependency Prediction.} 
We design a schema-dependency predictor to obtain the dependency between questions and schemas.
Here, we use the deep biaffine mechanism \cite{Dozat2017DeepBA,Dozat2018SimplerBM} in Eq. \ref{biaff}, which is a popular mechanism widely used in dependency parsing tasks.
It decomposes the dependency prediction into the presence or absence of dependency (edge), and the type of potential edge (label).

\begin{equation}\label{biaff}
\operatorname{biaff}\left(\mathbf{h}_{1}, \mathbf{h}_{2}\right)=\mathbf{h}_{1}^{\top} \mathbf{Uh}_{2}+\mathbf{W}\left(\mathbf{h}_{1} \oplus \mathbf{h}_{2}\right)+\mathbf{b}
\end{equation}
where the $\mathbf{U}, \mathbf{W}, \mathbf{b}$ is the learnable parameters.
Taking a question and schema ${[\mathbf{x}, \mathbf{h}]}$ as input, the schema dependency module first builds the unified representation $z$ of the input through a $\operatorname{Bi-LSTM}$ \cite{DBLP:journals/neco/HochreiterS97} in Eq. \ref{lstm}.
\begin{equation}\label{lstm}
\left(\mathbf{z}_{1}, \ldots, \mathbf{z}_{n+m}\right)=\operatorname{Bi-LSTM}\left(\mathbf{x}_{1}, \ldots, \mathbf{h}_{m}\right),
\end{equation}
Then, we use the single-layer feedforward network (FFN) to reduce the dimension and obtain the specific head and dependence representations in Eq. \ref{mlp}
\begin{equation}\label{mlp}
\begin{aligned}
\mathbf{r}_{i}^{(\text {edge-head })} &=\operatorname{FFN}_{\text {edge-head }}\left(\mathbf{z}_{i}\right), \\
\mathbf{r}_{i}^{(\text {label-head })} &=\operatorname{FFN}_{\text {label-head }}\left(\mathbf{z}_{i}\right), \\
\mathbf{r}_{i}^{(\text {edge-dep })} &=\operatorname{FFN}_{\text {edge-dep }}\left(\mathbf{z}_{i}\right), \\
\mathbf{r}_{i}^{(\text {label-dep })} &=\operatorname{FFN}_{\text {label-dep }}\left(\mathbf{z}_{i}\right).
\end{aligned}
\end{equation}
Next, we perform the biaffine attention mechanism to capture the complex dependency in Eq. \ref{biaff_forward}.
\begin{equation}\label{biaff_forward}
\begin{array}{l}
s_{i, j}^{(\text {edge })}=\operatorname{biaff}_{\text {edge}}\left(\mathbf{r}_{i}^{(\text {edge-dep })}, \mathbf{r}_{j}^{(\text {edge-head })}\right), \\
{s}_{i, j}^{(\text {label })}=\operatorname{biaff}_{\text {label}}\left(\mathbf{r}_{i}^{(\text {label-dep })}, \mathbf{r}_{j}^{\text {(label-head) }}\right), \\
y_{i, j}^{\prime(\text{edge})}=\left\{s_{i, j} \geq 0\right\}, \\
y_{i, j}^{\prime(\text{label})}=\operatorname{softmax} ({s}_{i, j}^{(\text {label })}).
\end{array}
\end{equation}
Finally, we optimize model as follows with the cross-entropy loss:
\begin{equation}
\resizebox{.85\hsize}{!}{$
\begin{aligned}
\mathcal{L}_{dep}&=\sum_{i=1}^{n+m} \sum_{j=1}^{n+m} (-y_{i,j}^{\text{edge}} \log y_{i, j}^{\prime(\text{edge})} - y_{i,j}^{label} \log y_{i, j}^{\prime(\text{label})})
\end{aligned}
$}
\end{equation}

\subsection{SQL Prediction}
We follows \cite{Hwang2019ACE} to build sketch-based SQL prediction module.
It consists of a series of sub-modules, each predicting a part of the final SQL independently.
Due to space limitations, please read Appendix A for a detailed description of the network.
Finally, we compute the standard cross-entropy loss $\mathcal{L}_{sql}$ which is the sum of the sub-module cross-entropy losses.

\subsection{Adaptive Multi-task Loss}
For the multi-task learning, the loss of the two sub-tasks can be integrated directly through weighting, but these weights depend on empirical setting.
In \texttt{SDSQL}, we use the adaptive loss \cite{Kendall2018MultitaskLU}, which learn a relative weighting automatically from the data.
The final loss function for \texttt{SDSQL} is:
\begin{equation}
\mathcal{L} = \frac{1}{2 \sigma_{1}^{2}} \mathcal{L}_{dep}+\frac{1}{2 \sigma_{2}^{2}} \mathcal{L}_{sql}+\log \sigma_{1} \sigma_{2}
\end{equation}
where the $\sigma_{1}$ and $\sigma_{2}$ are learnable parameters.

\section{Complex SQL Expansion}
We counted the cases in real business scenarios and found that over 85\% of them are simple SQL, and improvements made on them bear a significant impact on real-life applications.
For complex SQL, \eg, Spider dataset, the generation-based \citep{guo-etal-2019-towards,wang-etal-2020-rat} (IRNet, RATSQL) and sketch-based  approach \citep{Choi2020RYANSQLRA,Zeng2020RECPARSERAR} outperformance on it. The latter ones have slot prediction modules similar to SQLNet for the WikiSQL, while recursion  modules are  introduced  to  handle  the generation of nested SQL sketches, a characteristic in Spider but absent in WikiSQL. We are considering extending our method by existing sketch-based methods as in RYAN-SQL, while introducing our schema dependency methods on Spider.

\section{Experiment}

\subsection{Setup}

\paragraph{Dataset and Evaluation Metrics.}
WikiSQL \citep{Zhong2017Seq2SQLGS} is a collection of questions, corresponding SQL queries, and SQL tables from real-world data extracted from the web.
For evaluation, the logical form accuracy (LF) is the percentage of strict string matching between predicted SQL queries and labels; the execution accuracy (EX) is the percentage of exact matches of executed results of predicted SQL queries and labels.


\paragraph{Compared Models.}
We compare the proposed method to the following state-of-the-art models: (1) {Seq2SQL} \citep{Zhong2017Seq2SQLGS} is a generation-based baseline;   (2) {SQLnet} \citep{Xu2017SQLNetGS} is  a sketch-based  method;  (3) TypeSQL \citep{Xu2017SQLNetGS} utilizes  type  information to better understand rare entities and numbers in questions; (4) SQLova \citep{Hwang2019ACE} first integrates the pre-trained language model in the sketch-based method; (5) X-SQL \citep{He2019XSQLRS}  enhances the structural schema representation with the contextual embedding; (6) HydraNet \citep{Lyu2020HybridRN} breaks down the problem into column-wise ranking and decoding;
(7) IESQL \citep{Ma2020MentionEA} proposes an information extraction approach to Text-to-SQL and tackles the task via sequence-labeling-based relation extraction.
(8) RATSQL \citep{wang-etal-2020-rat} \footnote{the RATSQL with BERT model cannot converge on WikiSQL dataset, here reported the result of BERT-less model.} use the relation transformer layers for schema linking.

\begin{table}
	\centering
	\scalebox{0.9}{
	\begin{tabular}{ccccc}
		\toprule 
		{\multirow{3}*{\textbf{Model}}} & \multicolumn{2}{c}{Dev} & \multicolumn{2}{c}{Test} \\
		\cmidrule{2-5}
		& LF & EX & LF & EX  \\
		\midrule
		Seq2SQL & 49.5 & 60.8 & 48.3 & 59.4\\
		SQLNet & 63.2 & 69.8 & 61.3 & 68.0 \\
		TypeSQL & 68.0 & 74.5 & 66.7 & 73.5 \\
		RATSQL & 73.6 & 82.0 & 75.4 & 81.4 \\
		SQLova & 81.6 & 87.2 & 80.7 & 86.2 \\
		X-SQL & 83.8 & 89.5 & 83.3 & 88.7  \\
		HydraNet & 83.6 & 89.1 & 83.8 & 89.2  \\
		IESQL & 81.1 & 86.5 & 81.1 & 86.5 \\
		SDSQL & \textbf{86.0} & \textbf{91.8} & \textbf{85.6} & \textbf{91.4} \\
		\bottomrule
	\end{tabular}
	}
	\caption{Performance of various methods in both dev and test on WikiSQL dataset.}
	\vspace{-0.8cm}
    \label{result}
\end{table}

\subsection{Implementatation Details}
We utilize PyTorch \citep{DBLP:conf/nips/PaszkeGMLBCKLGA19} to implement our proposed model.
A natural language question is first tokenized by with CoreNLP \citep{manning-etal-2014-stanford} and further tokenized by WordPiece \cite{Devlin2019BERTPO}.
For the input representation, we use bert-large-uncased version \cite{Devlin2019BERTPO} and fine-turn it with 1e-5 learning rate during training.
We use Adam \citep{KingmaB14} to minimize loss and set the learning rate as 1e-3 for the SQL prediction module and a learning rate of 1e-4 for the schema dependency module.

\begin{table}[!htbp]
	\centering
	\scalebox{0.9}{
	\begin{tabular}{ccccc}
		\toprule 
		{\multirow{3}*{\textbf{Model}}} & \multicolumn{2}{c}{Dev} & \multicolumn{2}{c}{Test} \\
		\cmidrule{2-5}
		& LF & EX & LF & EX  \\
		\midrule
		SQLova + EG & 84.2 & 90.2 & 83.6 & 89.6 \\
		X-SQL + EG & 86.2 & 92.3 & 86.0 & 91.8  \\
		HydraNet + EG & 86.6 & 92.4 & 86.5 & 92.2  \\
		IESQL + EG & 85.8 & 91.6 & 85.6 & 91.2 \\
		SDSQL + EG & \textbf{86.7} & \textbf{92.5} & \textbf{86.6} & \textbf{92.4} \\
		\bottomrule
	\end{tabular}
	}
	\caption{Performance of various methods with execution guided (EG) decoding strategy.}
	\vspace{-0.5cm}
    \label{eg}
\end{table}

\subsection{Overall Performance}
We first compare the performance of \texttt{SDSQL} with other state-of-the-art models on the WikiSQL benchmark without using EG. As shown in Table~\ref{result}, we can see that \texttt{SDSQL} outperforms all existing models on all evaluation metrics.
To explore the impact of EG, we compare the performances of various methods with EG in Table~\ref{eg}.

\begin{figure}[!htbp]
	\centering
	\includegraphics[width=0.65\linewidth]{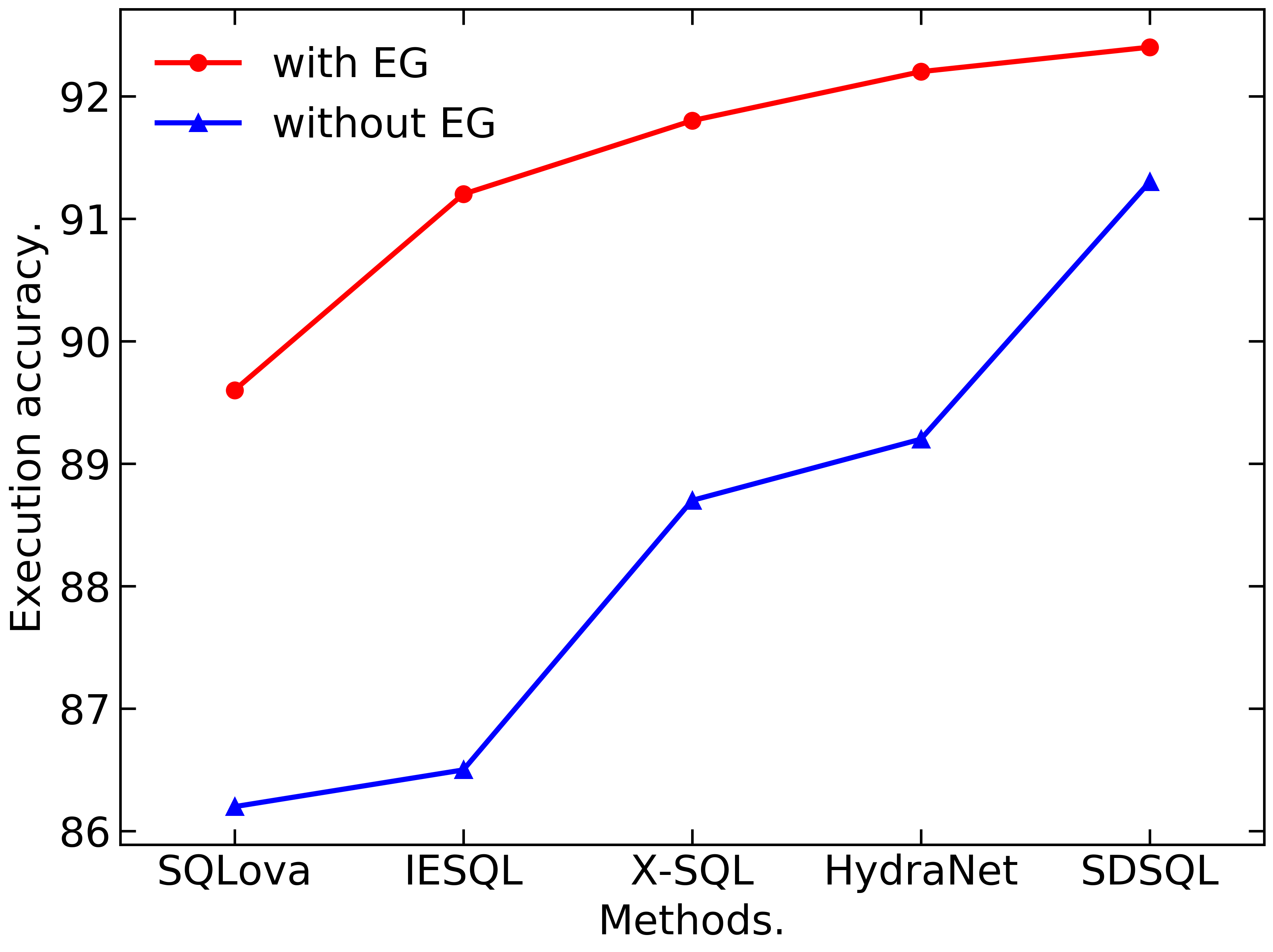}
	\caption{Comparison of methods with or without EG. }
	\label{eg_fig}
\end{figure}

We can see that our \texttt{SDSQL+EG} still achieves the best reported result on WikiSQL.
In Figure \ref{eg_fig}, we further show the impact of EG on different models. We can see that the additional benefit of using EG for \texttt{SDSQL} is minimum, suggesting that the proposed schema dependency method can cover some benefits of EG and alleviate the need for it.  
In general, the condition of triggering EG is that an error occurs during execution. This type of error is often caused by the illegal columns and values of the generated SQL. Schema dependency completes the linking task better, and the illegal proportion of the generated SQL is hence reduced. 

To further evaluate the impact of EG on inference time, we list the time consumption in Table \ref{time}. To emulate the real application scenario, we evaluate time consumption with CPU and a batch size of 1. We can see that the strategy without EG significantly reduces time consumption and improves inference efficiency, sacrificing only a small amount of performance and providing more flexibility for practical use.


\begin{table}[!htbp]
	\centering
	\scalebox{0.9}{
	\begin{tabular}{ccc}
		\toprule 
		Sample Num. & SDSQL & SDSQL + EG \\
		\midrule
		\#100K & 350772 (s) & 663218 (s)  \\
		\bottomrule
	\end{tabular}
	}
	\caption{Comparison of the inference time in seconds. }
	\vspace{-0.5cm}
    \label{time}
\end{table}

\subsection{Ablation Study}
To understand the importance of each part of \texttt{SDSQL}, we present an ablation study in Table \ref{ablation}.
\texttt{SDSQL} has two new components compared to base model (SQLova) \cite{Hwang2019ACE}, \ie, the {Schema dependency (SD) module and adaptive multi-task loss. }
Here we incrementally add the two components to observe the performance change. 
It is worth noting that the schema dependency task improves performance more significantly.

\begin{table}[!htbp]
	\centering
	\scalebox{0.9}{
	\begin{tabular}{lcccc}
		\toprule 
		{\multirow{3}*{\textbf{Model}}} & \multicolumn{2}{c}{Dev} & \multicolumn{2}{c}{Test} \\
		\cmidrule{2-5}
		& LF & EX & LF & EX  \\
		\midrule
		Base & 81.6 & 87.2 & 80.7 & 86.2\\
		    (+) SD module & 85.0 & 90.6 & 85.0 & 90.8 \\
		    (+) adaptive loss & 85.5 & 91.3 & 85.6 & 91.4\\
		\bottomrule
	\end{tabular}
	}
	\caption{Ablation study of \texttt{SDSQL} over LF and EX in both dev and test on WikiSQL dataset.}
	\vspace{-0.5cm}
    \label{ablation}
\end{table}

\subsection{Further Analysis}
In order to investigate the gain from schema dependency, we performed a fine-grained analysis as shown in \ref{fine}.
We observed that \texttt{SDSQL} is improved for all sub-modules, especially \textit{W-Col} and \textit{W-Val}.
The reason for our analysis is mainly due to schema dependency could capture the complex interaction and help model to link columns and values with their corresponding schema through dependencies.

\begin{table}[!htbp]
	\centering
	\scalebox{0.9}{
	\begin{tabular}{ccccccc}
		\toprule 
	    Method & $s_{col}$ & $s_{agg}$ & $w_{no}$ & $w_{col}$ & $w_{op}$ & $w_{val}$  \\
		\midrule
		SQLova & 96.8 & 90.6 & 98.5 & 94.3 & 97.3 & 95.4  \\
		SDSQL  & 97.3 & 90.9 & 98.8 & \textbf{98.1} & 97.7 & \textbf{98.3} \\
		\bottomrule
	\end{tabular}
	}
	\caption{Fine-grained analysis for each sub-module in SQL prediction task.}
    \label{fine}
\end{table}



\subsection{AGG Prediction Enhancement}
The AGG prediction is a bottleneck for Text-to-SQL model in wikiSQL, since the AGG
annotations in dataset have up to 10\% of errors \citep{Hwang2019ACE}.
Following IESQL, we add some rules based on the word tuple co-occurrence features as the AGG Prediction Enhancement (AE).
It should be emphasized that AE is equivalent to \textbf{fitting the flawed annotations}, and is not necessary for really Text-to-SQL task.
We add AGG enhancement here only for fair comparison with IESQL. As shown in Table \ref{AE}, after adding AE operation, the execution accuracy of SDSQL is outperform the IESQL.

\begin{table}[!htbp]
	\centering
	\scalebox{0.9}{
	\begin{tabular}{ccc}
		\toprule 
	    Modle & LF & EX  \\
		\midrule
		IESQL + AE & 87.8 & 92.5  \\
		SDSQL + AE & 87.0 & \textbf{92.7} \\
		\bottomrule
	\end{tabular}
	}
	\caption{Comparison of the IESQL and SDSQL with agg prediction enhancement performance.}
	\vspace{-0.5cm}
    \label{AE}
\end{table}


\section{Conclusions}
This paper proposes a novel multi-task Text-to-SQL model that integrates schema dependency to capture the complex interaction between schemas and questions.
The proposed \texttt{SDSQL} model outperforms all existing models on the WikiSQL benchmark. In the setup without the EG strategy, it significantly speeds up the inference without  sacrificing much performance, providing  flexibility for supporting practical applications.

\bibliography{anthology,custom}
\bibliographystyle{acl_natbib}

\clearpage

\appendix

\section{SQL Prediction Module Details} \label{sqlova}
The SQL prediction module is follow with \cite{Hwang2019ACE} and descried here for comprehensive reading.
It consists a series of sub-modules that independently predict each part of SQL.
In each sub-module, column-attention is applied to reflect the most relevant information in natural language questions when prediction is made on a particular column:

\begin{equation}
\begin{aligned}
\alpha &=\operatorname{softmax}({H}^T \mathbf{W}_{att} {X}) \\
C &=\sum_{n} \alpha_n \times x_n
\end{aligned}
\end{equation}
where the $\mathbf{W}_{att}$ is the learnable parameters, $C$ is the embedding of question across the schema headers.
A complete SQL can be decompose into the \textit{Select} Clause and \textit{Where} Clause.
\paragraph{Select Clause.} The \textit{S-Col} module aims to find the column according to the question and schema, and \textit{S-Agg} module aims finds aggregation operator.
\begin{equation}
p_{sc} =\operatorname{softmax}(\mathbf{W}_{sc} \tanh ([\mathbf{U}_{sc}^{h} H;\mathbf{U}_{sc}^{q} C]))
\end{equation}
where the $\mathbf{W}, \mathbf{U}$ are learnable matrices.
The \textit{S-Agg} module finds aggregation operator for six possible choices: {\{\textit{None}, \textit{Max}, \textit{Min}, \textit{Count}, \textit{Sum}, \textit{Avg}\}}.
\vspace{-0.5cm}

\begin{equation}
p_{sa} = \operatorname{softmax}\left(\mathbf{W}_{sa} \tanh ( \mathbf{U}_{sa}^{q} C)\right)
\end{equation}

\paragraph{Where Clause.} 
The first component in this part is \textit{W-Num}. It predicts the number of column in the \textit{where} clause as the $(k + 1)$ way classification model, where $k$ is the max column number. 

\begin{equation}
\begin{array}{c}
C_{h} = \operatorname{SA} (\mathbf{H}) \\
C_{h}^{h} = \mathbf{W}_{h} C_{h} \\
C_{h}^{c} = \mathbf{W}_{c} C_{h} \\
C_{x} = \operatorname{softmax} (\mathbf{W} \operatorname{Bi-LSTM_{wn}}([X, C_{h}^{h}, C_{h}^{c}])) \\
p_{wn} = \operatorname{softmax}\left(\mathbf{W}_{wn} \tanh ( \mathbf{U}_{wn}^{q} C_{x})\right) \\
\end{array}
\end{equation}
where the $\operatorname{SA}$ is the self-attention mechanism \cite{NIPS2017_3f5ee243} to capture the internal relation of the schema, and $C_{h}^{h}$ and $C_{h}^{c}$ are initial the $\operatorname{Bi-LSTM_{wn}}$ input, \ie, \textit{hidden} and \textit{cell} vector. 
Similar to \textit{S-Col}, the \textit{W-Col} module predicts column through the column attention vector:
\begin{equation}
p_{wc} =\sigma(\mathbf{W}_{wc} \tanh ([\mathbf{U}_{wc}^{h} H;\mathbf{U}_{wc}^{q} C]))
\end{equation}
where $\sigma$ is the $\operatorname{sigmoid}$ function, which obtains the probability of selection in the top $k$ column.
Further more, the \textit{W-op} has three choices: \{\textit{=}, \textit{>}, \textit{<}\} and \textit{W-val} finds where condition by locating the starting and ending tokens from the question for the given column and operator.

\begin{equation}
\begin{array}{c}
p_{w o}=\operatorname{softmax}\left(\mathbf{W}_{w o} \tanh \left(\left[\mathbf{U}_{w o}^{h} H ; \mathbf{U}_{w o}^{q} C\right]\right)\right) \\
p_{w v}=\left(\mathbf{W}_{w v} \tanh \left(\left[X ; \mathbf{U}_{w v}^{h} H ; \mathbf{U}_{w v}^{q} C ; \mathbf{U}_{w v}^{o p} V\right]\right)\right)
\end{array}
\end{equation}
where $V$ is the one-hot vector for indicating operator.
Finally, we compute the standard cross-entropy loss $\mathcal{L}_{sql}$ which is the sum of the sub-module cross-entropy losses.

\end{document}